\documentclass[fleqn,10pt,twocolumn]{AROB-ISBC-SWARM22}
\usepackage{amsmath}
\usepackage{algorithm} 
\usepackage{algpseudocode} 
\usepackage{array}
\usepackage{multirow}

\title{Message Expiration-Based Distributed Multi-Robot Task Management}

\author{Yikang Gui, Ehsan Latif, Ramviyas Parasuraman}
\speaker{Yikang Gui}
\affils{Department of Computer Science, University of Georgia, Athens, GA 30602, USA\\
(Corresponding author email: ramviyas@uga.edu)}
\abstract{%
Distributed task assignment for multiple agents raises fundamental and novel control theory and robotics problems. A new challenge is the development of distributed algorithms that dynamically assign tasks to multiple agents, not relying on prior assignment information. This work presents a distributed method for multi-robot task management based on a message expiration-based validation approach. Our approach handles the conflicts caused by a disconnection in the distributed multi-robot system by using distance-based and timestamp-based measurements to validate the task allocation for each robot. Simulation experiments in the Robotarium simulator platform have verified the validity of the proposed approach.}
\keywords{%
Distributed System, Multi-Robot, Task Allocation, Message Expiration}
\begin{document}

\maketitle


\section{Introduction}

Since the early 1990s, the problem of task allocation in multi-robot systems has received significant, and increasing interest in the research community \cite{khamis2015multi}. As researchers design, build and use cooperative multi-robot systems, they invariably encounter the question: "Which robot should execute which task?" This question must be answered, even for relatively simple multi-robot systems, and the importance of task allocation grows with the complexity, size, and capability of the system under study. Even in the simplest case of homogeneous robots with fixed, identical roles, intelligent allocation of tasks is required for good system performance, if only to minimize physical interference.

Recent advances in communication and computation have given rise to distributed control of multiagent systems which, compared to conventional centralized control, provide increased efficiency, performance, and scalability as well as
robustness due to its ability to adjust to agent failures or dynamically changing environments \cite{lenka2018,michael2008distributed,firoozi2020distributed}. Inspired by these appealing properties of distributed control, we propose a distributed and online solution to the multiagent dynamic task allocation problem, where the assignment of robots to tasks may need to be continuously adjusted depending on changes in the task environment.

This paper focuses on multi-robot task management to achieve high performance and efficiency in sample foraging tasks with multiple homogeneous (swarm) robots. Additionally, we are dealing with a distributed multi-robot system instead of a centralized multi-robot system to improve the scalability of the proposed method. 

Most multi-robot task allocation algorithms focus on the efficiency of task accomplishment based on strict assumptions. The assumptions include that the energy level of the robot will not be a problem or that the communication of the robots is always stable. However, in practice, there are many constraints on the robots, especially for the swarm that they are designed in a small shape, which leads to poor communication signals and limited battery \cite{parasuraman2012energy}. Similarly, swarm robots cannot have a sufficient battery to finish multiple tasks without recharging due to physical constraints. With an excellent recharging algorithm, the swarm robots can work on their jobs automatically. On the other hand, heterogeneity in the robot's capability such as their max speed, energy level, variations on functional intelligence, etc. can also impact the task performance \cite{ov2020impact}.

Yang et al. \cite{yang2020needs,yang2019self} introduced a novel framework for cooperation between heterogeneous multi-robot systems for dynamic task assignments and automated planning. It combines robot perception, communication, planning, and execution in MRS, which considers individual robots' needs and action plans and emphasizes the complex relationships created through communication between the robots. Specifically, they proposed Robot's Needs Hierarchy to model the robot's motivation and offer priority-based task management. In \cite{djenadi2021energy}, the authors presented a distributed energy management task allocation strategy for multi-robot systems involved in goods transportation, considering energy management in the task allocation. Authors in \cite{fouad2020energy} presented a control barrier function-based framework for long-term autonomy of multi-robot systems with limited charging resources. 

Motivated by these two demands on energy and communication for swarm robots, we propose a novel distributed algorithm that can handle the disconnectivity and the recharging problem and achieve continuous performance in repetitive tasks. Specifically, this paper presents a multi-robot task management algorithm integrative of energy-awareness and connectivity-awareness focused on distributed tasks by introducing a new expiration-based validation mechanism to avoid conflicts in task allocation.

\section{PROBLEM FORMULATION}
Let us denote $R = \{R_1, R_2, ..., R_n\}$ as the multi-robot systems including $n$ Robots. For each robot $R_i$, it will have its neighbors, such as $N_i = \{R_1, R_3, R_5\}$. What is more, for each robot $R_i$, it will have its own task schedule including other robots information, denoted as $S_i=\{S_i^1, S_i^2, ..., S_i^n\}$. 

The problem is to synchronize the task schedule over each robot with the following constraints: the connectivity may vary due to the sensing range. Because of the unstable connectivity, one robot may not be able to synchronize with all other robots.
\begin{equation}
\begin{cases}
  \text{Synchronize}(S_i, S_j), & \text{if $S_i$ is connected with $S_j$}
\\
  \text{Do nothing}, & \text{otherwise}
\end{cases}
\end{equation}

\section{Proposed Methodology}

The proposed task management algorithm involves three main components: 1) task receipt handler; 2) robot task scheduler; 3) robot task controller.

\subsection{Task Receipt}
A task receipt is a message containing the status information of a robot. The content includes:
\begin{itemize}
    \item Task type: Currently, there are 7 task types, including IDLE, GO\_TO\_PICK, GO\_TO\_COLLECTION, GO\_TO\_RECHARGE, RECHARGING, \\WAIT\_FOR\_RECHARGING, DEAD.
    \item Task location: The coordinates of the current task
    \item Robot id: The robot id for this message
    \item Previous task location: The coordinates of the previous one task, only useful when the current task is GO\_TO\_COLLECTION
    \item Timestamp: Current global timestamp
    \item Priority: Useful for validation of this message
    \item Distance to task location: The Euclidean distance to the current task
\end{itemize}

\subsection{Robot Task Scheduler}
In the \textit{RobotStatusManager}, there is a \textit{Task Controller} to control the task schedule stored in every robot. The task schedule stores the information about itself and all other robots. The information about other robots may be outdated due to disconnectivity.

\subsection{Robot Task Controller}
A \textit{TaskController} is used for managing all task receipts of one robot. The number of task receipts should be the same as the number of robots. A fully functional \textit{TaskController} should return the status of every task in the environment for future purposes, such as task assignments. 

~\
\subsubsection{Check Connectivity}
Since in our setting, every robot has the same sensing range. The robot can communicate with its neighbors only when its neighbors are in its sensing range. It is very important for the robot to determine whether its neighbors are connected or not. In order to achieve this, we define an algorithm named Check Connectivity. The procedure is illustrated in Algorithm \ref{cc}.

\begin{algorithm}[h]
	\caption{Check Connectivity} 
    \label{cc}
	\begin{algorithmic}[1]
	    \State Calculate the distance between each pair of robots
	    \State Keep the pair of robots whose distance between them is smaller than the sensing range
	    \State Save the above pair of robots for future purpose
	\end{algorithmic}
\end{algorithm}

\subsubsection{Check Expiration}
In order to avoid the scenario that one robot was broadcast with all other robots in the beginning, and it got disconnected in the rest simulation. In that case, the robot will never take a movement, instead of waiting for reconnected forever. Therefore, we introduce a mechanism called \textit{expiration}. The task receipts saved in each robot task schedule will have an expiration. Once the task receipts is expired, the robot task schedule will assign an idle task receipt to the targeted robot. Here is a difference in the task receipt assigned after expiration that the priority will assign to 0 instead of 1. Priority 1 means that this task receipt is assigned by its real owner. Priority 0 means that the task receipt is assigned after expiration, and its real owner does not assign it. Here, the owner means that the robot with the \textit{robot ID} in the task receipt. 

Here is an example. Assume that there are two robots, robot A and robot B, in our simulation. The expiration threshold is 400, and there is only one treasure location. Robot A and robot B are initialized close enough to make sure they are connected. Robot A is assigned to take the treasure, and Robot B is assigned idle at timestamp 0. Due to the connectivity, robot A and robot B have the same task schedule, in which robot A is recorded to take the treasure and robot B is recorded to be idle. Then robot A goes to fetch the treasure, and robot A and robot B get disconnected in the rest simulation. At timestamp 401, robot B will update the status of robot A in its task schedule with idle due to the expiration of robot A's task receipt. What's more, since the task receipt of robot A in robot B's \textit{TaskController} is updated by robot B, the priority of this task receipt is 0. Algorithm \ref{expiration} depicts the procedure to fix the expiration.

\begin{algorithm}[ht]
	\caption{Check Expiration} 
	\label{expiration}
	\begin{algorithmic}[1]
	    \State EXPIRATION is the threshold for expiration
	    \For {Every \textit{TaskReceipt} in task schedule}
    	    \If {\textit{TaskReceipt}.timestamp - current\_timestamp $>$ EXPIRATION}
    	        \If {\textit{TaskReceipt}.priority $>$ 0}
    	            \If {\textit{TaskReceipt}.robotID != its own robotID}
    	                \State \textit{TaskReceipt} = IDLE \textit{TaskReceipt} with priority = 0
                    \EndIf
                \EndIf
            \EndIf
	    \EndFor
    \State \textbf{return} task schedule
	\end{algorithmic}
\end{algorithm}

\subsubsection{Check Validation}
This function is designed to handle the conflicts caused by the disconnectivity of each pair of robots. 

The setting of the environment causes the first set of conflicts, the conflicts happen when two or more robots are assigned the same treasure, and they are to go to pick the treasure. In this case, the robot closest to the treasure will keep the task, and the rest robots will  re-assign an idle task.

The second set of conflicts is all the rest of the possible conflicts. We can use timestamp and priority to handle this set of conflicts. When conflicts happen, we first check the priority of both task receipts. The task receipt with lower priority will be replaced by the task receipt with higher priority. If two task receipts have the same priority, the task receipts with the latest timestamp with replacing the other task receipt. Table \ref{measurement table} demonstrates the measurements to handle the conflicts. Algorithm \ref{validation} conveys the procedure to handle the conflicts.

\begin{algorithm}
	\caption{Check Validation} 
	\label{validation}
	\begin{algorithmic}[1]
        \For {TaskReceiptA, TaskReceiptB in TaskScheduleA, TaskScheduleB}
            \If {TaskReceiptA.priority $>$ TaskReceiptB.priority}
                \State TaskReceiptB = TaskReceiptA
            \ElsIf{TaskReceiptA.priority $<$ TaskReceiptB.priority}
                \State TaskReceiptA = TaskReceiptB
            \ElsIf{TaskReceiptA.timestamp $>$ TaskReceiptB.timestamp}
                
            \ElsIf{TaskReceiptA.timestamp $<$ TaskReceiptB.timestamp}
                \State TaskReceiptA = TaskReceiptB
            \EndIf
        \EndFor
        \While{multiple task receipts have PICK task and have same treasure\\}{
        keep the robot closest to the treasure and reassign IDLE to rest robots
        }
        \EndWhile
	\end{algorithmic}
\end{algorithm}

The below table (Table~\ref{measurement table}) presents how the validation is conducted between two robots for two different tasks (Pick treasure and go to collection).
\begin{table}[ht]
\caption{Measurements used for checking Validation.}
\label{measurement table}
\resizebox{\columnwidth}{!}{
\begin{tabular}{|l|l|l|}
\hline
Robot A Action & Robot B Action & Measurement for validation            \\ \hline
Pick           & Pick           & Distance               \\ \hline
Pick           & Collection     & Timestamp and priority \\ \hline
Collection     & Collection     & Timestamp and priority \\ \hline
Pick           & Idle           & Timestamp and priority \\ \hline
...            & ...            & Timestamp and priority \\ \hline
\end{tabular}}
\end{table}

\section{Simulation Experiments}
To verify the effectiveness of the proposed approach, we used the simulator version of the Robotarium platform \cite{pickem2016}, which is a multi-robot testbed. In our simulations, the following parameters are used.
\begin{itemize}
    \item N=5 number of robots.
    \item T=5 fixed treasure locations.
    \item R=2 fixed recharger stations at the top right and top left of the workspace.
    \item C=1 one fixed collection point at the bottom left of the workspace.
    \item I=10000 iterations for simulation
    \item Starting position of the robots = random within the workspace.
    \item Distance clearance threshold = 0.1m (to decide if a robot has reached a location or not).
    \item All distances are expressed in the Euclidean norm, and time is expressed as iterations.
    \item Disable the collision avoidance and barrier function in Robotarium.
\end{itemize}

\begin{figure}[t]
\centering
\includegraphics[width=\columnwidth]{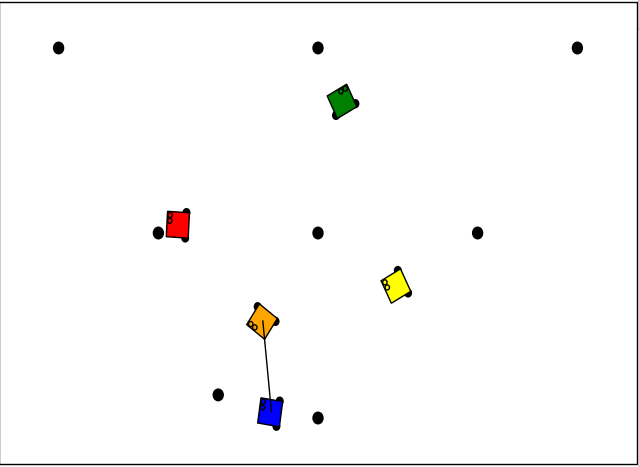}
\caption{Robotarium environment used in the experiments.}
\label{env}
\end{figure}

Fig.~\ref{env} shows an example of the simulation environment. The line between two robots represents they are connected. The black dots represent treasure, recharge station, and collection point.

\textbf{Assumption}: Every task location has an unlimited amount of treasure available. When a robot is at the task location, it is assumed that it picked up the treasure (if it does not have one already). When a robot reaches a collection point, then it is assumed that it dropped the treasure (if it already had one with it).

\textbf{Definition of Task completion}: A task is said to be complete when a robot reaches any of the task locations, and it had then dropped that treasure at the collection point.

\textbf{Energy model}:
A robot will lose energy based on the following energy consumption model \cite{parasuraman2014model}:
\begin{equation}
    \centering
    E_t = E_{t-1} - (\alpha \times 1 + \beta \times distance(t) + \gamma \times \text{isTreasurePicked}),
\end{equation}
where $\alpha$ represents the coefficient for the static energy consumption due to onboard computing and sensors of the robot for example, $\beta$ represents the coefficient for the dynamic power consumption by the robot due to movement (e.g, velocity-based motor power to move to a specific distance), and $\gamma$ represents the coefficient for the energy spent in picking up the treasure (e.g., manipulation effort).

Similarly, a robot will gain energy if it is at the recharging station with the following model:
\begin{equation}
    \centering
E_t = E_{t-1} +\delta\times 1,
\end{equation}
where $\delta$ represents the rate at which the battery can be recharged. 
We use the following values: $\alpha = 0.01, \beta=0.1, \gamma=0.02,\delta =0.1$.

\begin{table*}[t]
\caption{Performance results in the Robotarium simulations.}
\label{tab:results}
\vspace{-2mm}
\resizebox{\textwidth}{!}{
\begin{tabular}{|l|l|l|l|l|l|l|l|l|l|l|l|l|}
\hline
\textbf{Performance Metric} & \multicolumn{2}{c|}{Experiment 1} & \multicolumn{2}{c|}{Experiment 2} & \multicolumn{2}{c|}{Experiment 3} & \multicolumn{2}{c|}{Experiment 4} & \multicolumn{2}{c|}{Experiment 5} & \multicolumn{2}{c|}{Experiment 6}  \\
\hline
& Mean    & Std    & Mean    & Std    & Mean    & Std    & Mean    & Std    & Mean    & Std    & Mean    & Std    \\ \hline
Travel Distance (m)       & 223.71  & 2.41   &227.35 &2.36  &225.96 &2.44 &222.07 &2.05 &197.4 &3.83 & 182.44 &8.77 \\ \hline
Go to recharge time (s)    & 5009.15 & 157.74 &5019.25 &174.24 &4969.88 &176.16 &5061.1 &163.69 &4482.45 &125.65 &4642.31 &173.24 \\ \hline
Wait for recharge time (s) & 1785.4  & 360.20 &1848.2 &343.65 &1825.3 &300.58 &1820.55 &206.96 &1420.9 &310.76 &1476.73 &395.65 \\ \hline
Recharge time (s)          & 8517.9  & 245.76 &8606.0 &186.35 &8502.45 &163.33 &8567.35 &159.22 &8048.4 &188.27 &7979.42 &210.46 \\ \hline
Treasure completion (\#)    & 97.55   & 1.74   &95.65 &1.71 &96.42 &1.57 &95.75 &1.54 &78.0 &1.67 &71.42 &3.67 \\ \hline
\end{tabular}}
\end{table*}

\subsection{Experiments conducted}
We conducted the following six experiment cases (each with 20 trials) to verify the proposed approach.
\begin{description}
    \item Experiment 1: T=5, Enable expiration, sensing range = 0.5
    \item Experiment 2: T=5, Enable expiration, sensing range = 5
    \item Experiment 3: T=5, Disable expiration, sensing range = 0.5
    \item Experiment 4: T=5, Disable expiration, sensing range = 5
    \item Experiment 5: T=2, Enable expiration, sensing range = 0.5
    \item Experiment 6: T=2, Disable expiration, sensing range = 0.5
\end{description}

While the first four experiments analyze the performance variations due to the sensing range and expiration mechanism, the last two experiments analyze the impact of expiration on a tighter resource constraint on the number of treasure locations.
The performance metrics in our simulations are the following: distance traveled collectively by the robots, time spent in going to recharge, time spent in waiting for recharge, time spent during recharging, and the number of treasures picked up.

\subsection{Results and Discussion}
Table~\ref{tab:results} presents the results for all the six experiment cases. It presents the mean and standard deviation (std) for each performance metric for 20 trials in each case.
The expiration mechanism is enabled in Experiments 1, 2, and 5.
It can be seen that for a smaller sensing range, s=0.5, enabling expiration improved the performance in terms of efficiency (less distance traveled) and performance (more treasures collected) for both T=2 and T=5 cases.
However, when the sensing range is too high (s=5), there was no difference in the metrics due to the expiration mechanism.
While the results show promising value to the proposed approach, more in-depth analyses need to be conducted to integrate our task management approach with collision avoidance strategy and with the existing methods from the literature on energy-aware task allocation algorithms.

\section{Conclusion}
We proposed a novel expiration-validation-based approach for multi-robot task management of persistent tasks without conflicts on an energy-constrained swarm robotics system. The proposed expiration mechanism improved the performance when the sensing range of the robots was small. 

\bibliography{references}
\bibliographystyle{IEEEtran}

\end{document}